\documentclass{article}

\usepackage{microtype}
\usepackage{graphicx}
\usepackage{subcaption}
\usepackage{booktabs}
\usepackage{hyperref}

\usepackage[preprint]{icml2026}

\usepackage{amsmath}
\usepackage{amssymb}
\usepackage{mathtools}
\usepackage{amsthm}

\usepackage[capitalize,noabbrev]{cleveref}

\theoremstyle{plain}

\theoremstyle{definition}

\theoremstyle{remark}

\usepackage[textsize=tiny]{todonotes}

\newcommand{\accsrc}{\mathrm{acc}_{\mathrm{src}}}

\icmltitlerunning{Separators Make Carry Propagation Learnable: The Geometry of Latent Carry in a Multiplication Transformer.}

\begin{document}

\twocolumn[
  \icmltitle{Separators Make Carry Propagation Learnable:\\The Geometry of Latent Carry in a Multiplication Transformer.}

\begin{icmlauthorlist}
  \icmlauthor{Sama Satariyan}{SCAI}
  \icmlauthor{Raphael Cousin}{SCAI}
  \icmlauthor{Gérard Biau}{LPSM,IUF,MEGA}
\end{icmlauthorlist}

\icmlaffiliation{SCAI}{Sorbonne Center for Artificial Intelligence (SCAI), Sorbonne Université, Paris, France}
\icmlaffiliation{LPSM}{Laboratoire de Probabilités, Statistique et Modélisation (LPSM), Sorbonne Université, CNRS, Paris, France}
\icmlaffiliation{IUF}{Institut universitaire de France (IUF), Paris, France}
\icmlaffiliation{MEGA}{MEGAVOLT team, Inria Paris, France}

\icmlcorrespondingauthor{Sama Satariyan}{sama.satariyan@etu.sorbonne-universite.fr}

  % \icmlkeywords{Machine Learning, ICML}

  \vskip 0.3in
]

\printAffiliationsAndNotice{}

\begin{abstract}
Transformers asked to multiply multi-digit numbers in a single forward
pass often fail, and interpretability studies of pretrained language
models find arithmetic solved by input-range heuristics rather than by
an explicit carry. We train small Llama-style transformers from
scratch on $4 \times 4$ multiplication without chain of thought and
find that the input format is decisive: inserting a space token
between digits raises exact-match accuracy from 1\% to 89\%. Output
positions are learned in carry-chain order, with the middle digits,
which have the longest-range dependencies, learned last. Inside the
model, the separator token that predicts each digit (its
\emph{prediction slot}) encodes the carry-in as an angle on a ring in
the residual stream; examples with more distinct carry values fill
more of the ring. Activation patching between examples matched on the
column sum shows that this state is causally used before the last
layer: patching the prediction slot alone transfers the source carry
in up to 84\% of cases after block~4 for one middle column of our best
model, while for other columns the carry is first assembled at the
neighboring answer slot before reaching its own. Remaining errors are
almost always off by one, consistent with a small error on the carry
or on the circular digit code.
\end{abstract}

% =====================================================
\section{Introduction}
\label{sec:intro}

Multi-digit arithmetic is a useful testbed for studying what
transformers learn: each problem has a verifiable answer, and failure
modes are diagnostic. Small-operand arithmetic is solved easily, but
multi-digit multiplication remains brittle even for frontier models
\citep{saxton2019analysing, bai2025multiplication}. The asymmetry with
addition points to long-range carry propagation: each output digit
depends on partial products from many input pairs, and the carries
themselves can be much larger than one and span several columns.

Recent interpretability work suggests that transformers do not acquire
a genuine carry-propagation mechanism. \citet{nikankin2024arithmetic}
find that pretrained LLMs solve arithmetic through input-range MLP
heuristics, with no subcircuit dedicated to carry routing;
\citet{deng2024dissecting} report that, in small transformers trained
on multiplication, the loss component associated with multi-digit
carry propagation converges last and remains noisiest throughout
training.

We revisit this picture by training small Llama-style transformers
from scratch on $4\!\times\!4$ multiplication, without chain of
thought, and ask two questions: \emph{what makes the carry learnable
in a single forward pass}, and \emph{when a model succeeds, how is the
carry represented and used?}

\begin{figure*}[h]
  \centering
  \includegraphics[width=2.1\columnwidth]{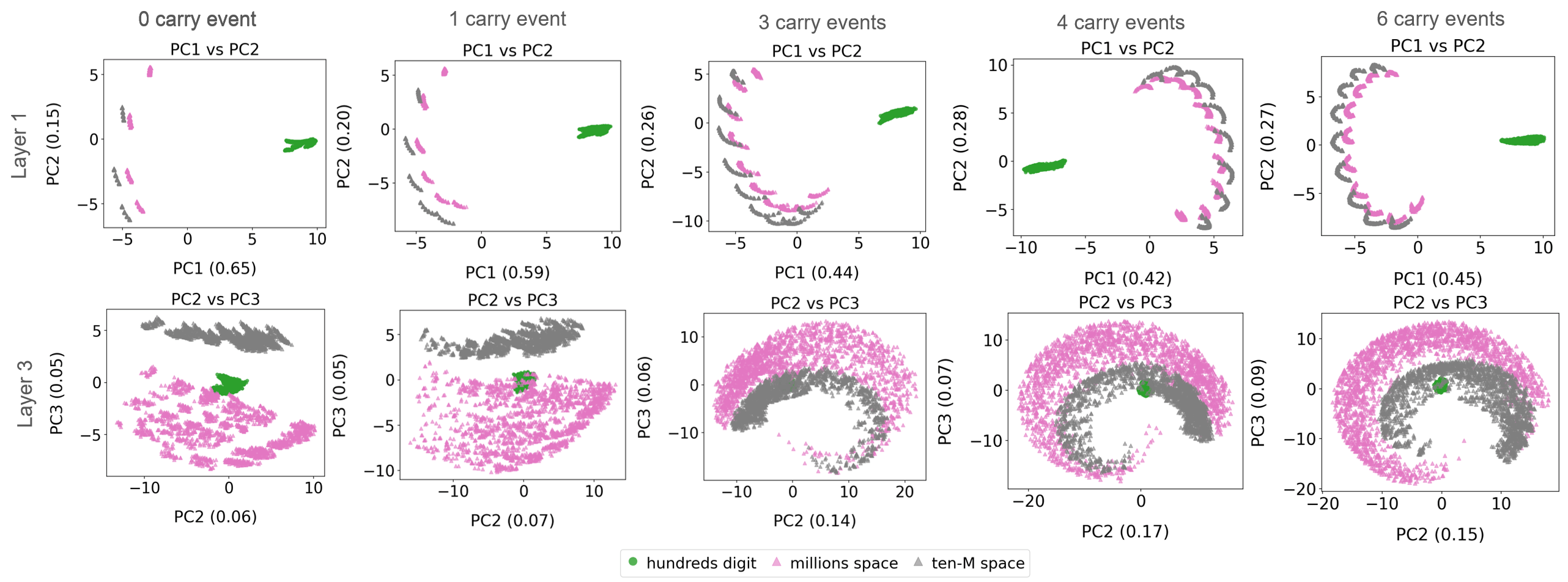}
  \caption{PCA of the residual stream at layers 1 and 3 for answer
  tokens, projected onto the first principal components. Digit and
  space tokens are colored by output position (see legend). Each
  panel corresponds to a different carry count $k$ (0, 1, 3, 4, 6),
  drawn from 2000 $4\times4$ multiplication examples. As $k$ grows,
  the space tokens trace a larger part of a ring: examples with more
  carries take more distinct carry values and so cover more of it.}
  \label{fig:mainpicture}
\end{figure*}

Our contributions are:
(i)~the format matters: with a space token between digits, a
4.8M-parameter model reaches 89\% exact match, and 1\% without it;
positions are learned in carry-chain order and the middle columns
never converge in the compact format;
(ii)~in the models that learn the task, the carry-in of each column is
encoded as an angle on a ring at the position that predicts its digit;
(iii)~matched activation patching shows this state is causally used at
intermediate layers, and that for  columns 4 and  5, the carry is first
assembled at the neighboring answer slot;
(iv)~residual errors at convergence are almost always $\pm 1$.

% =====================================================
\section{Related work}
\label{sec:related}

\paragraph{Mechanistic accounts of arithmetic.}
Prior work has reverse-engineered Fourier-basis digit representations
in modular addition \citep{nanda2023progress}, named carry and borrow
subcircuits in $n$-digit addition and subtraction
\citep{quirke2024understanding, quirke2024addsub}, and pentagonal-prism
digit geometry with Minkowski-sum-like combination in multiplication
\citep{bai2025multiplication}. A complementary line argues that
pretrained LLMs do not acquire algorithmic arithmetic at all, finding
only input-range MLP heuristics and behavioural patterns inconsistent
with a right-to-left carry algorithm \citep{nikankin2024arithmetic,
spilger2025lookahead, gambardella2024hard, yan2025grasp}. These
results are not in contradiction with ours: they concern large
pretrained models whose data never forced a systematic carry. We show
that a small model trained on multiplication, given a suitable format,
does build an explicit carry-in representation.

\paragraph{Geometry of latent space.}
Semantic content in transformer activations often takes specific
geometric forms: linear directions for truth, refusal, and
pre-committed answers \citep{marks2023geometry, arditi2024refusal,
cox2025decoding}; circular and helical manifolds for cyclic concepts
and digits modulo ten \citep{engels2024notalllinear,
kantamneni2025trigonometry}; orthogonal subspaces for hierarchies
\citep{park2024geometry}; linear axes for space and time
\citep{gurnee2023spacetime}; and fractal belief-state geometries for
Hidden Markov data \citep{shai2024belief}. Our finding adds a case in
which the shape encodes an intermediate computational quantity, the
carry, rather than a concept of the input.

\paragraph{Separator and filler tokens.}
\citet{pfau2024letsthink} show that filler tokens such as repeated
dots can host hidden computation; \citet{goyal2024pause,
pause2025expressivity} extend this to learned pause tokens. In our
setting the space token plays a related role: it adds a position with
no digit content before each output digit, and the model uses it as
the slot from which that digit is predicted and where its carry-in is
encoded.

% =====================================================
\section{Setup}
\label{sec:setup}

\paragraph{Model and data.}
We train a Llama-style decoder transformer with RMSNorm, RoPE, SwiGLU
and tied embeddings. The model has hidden size 256, 4 attention heads,
intermediate size 704, and 6 layers (4.8M non-embedding parameters).
We use 1M training examples of $4 \times 4$ multiplication with
operands in $[0, 9999]$, zero-padded to four digits, and answers
zero-padded to eight digits. Under uniform sampling, examples with
many carries are rare, so we skew the training distribution so that
every carry count $k \in \{0, \dots, 7\}$ (the number of output
columns receiving a non-zero carry) is well represented. In the
\emph{spaced} format a space token separates consecutive digits, in
MSB-first order, e.g.\
\texttt{0 1 2 3 * 0 4 5 6 = 0 0 0 5 6 0 8 8};
the \emph{compact} format removes the spaces. The model outputs the
answer directly, without chain of thought: with CoT it reaches 100\%
in 13k steps, but the computation is then written in the output and
there is nothing left to find inside the network.

For output column $c$ (LSB-indexed), with column sum
$S_c = \sum_{i+j=c} a_i b_j$ and carry-in $\gamma_c$,
\begin{equation}
  d_c = (S_c + \gamma_c) \bmod 10, \qquad
  \gamma_{c+1} = \left\lfloor \frac{S_c + \gamma_c}{10} \right\rfloor .
  \label{eq:carry}
\end{equation}
The carry-in reaches 30 in our data. Because the model predicts each
token from the position before it, the digit of column $c$ is read out
from the hidden state of the space token just before that digit. We
call this position the \emph{prediction slot} of column $c$.

\paragraph{Training and evaluation.}
We use AdamW (\texttt{lr=3e-4}, weight decay 0.1) with a cosine
schedule and 200 warmup steps, batch size 128, and 100K training
steps. Weights are initialised $\mathcal{N}(0, 0.02^2)$. The main
(spaced, 4-head) configuration is trained with three seeds, which
reach 89\%, 80\% and 78\% exact match. We evaluate on 2K held-out
examples and report exact-match accuracy and the per-position log
error $\log(|p_i - g_i| + 1)$, where $p_i$ and $g_i$ are the
predicted and ground-truth digits at output position $i$.

\paragraph{Hidden-state analysis.}
We record the residual stream $h^{(\ell)}(p)$ after every block
$\ell \in \{0, \dots, 5\}$ and at every token position $p$ (the
embedding plus the outputs of all attention and MLP blocks up to and
including block $\ell$). Within each (layer, position) group we fit a
PCA across examples; for curved structure we also use diffusion maps.
To probe the dependence on carry structure, we form groups of examples
with fixed carry count $k$ and fit separate PCAs per group. All
representational analysis uses the  seed 10.

\paragraph{Activation patching.}
We partition examples into a source group (carry-in $\gamma_c = c_s$
at a target output column) and a target group ($\gamma_c = c_t \neq
c_s$). For each target example we select the individual source
example whose column sum $S_c$ is closest (an exact match in 90--96\%
of cases across the columns tested), which holds fixed the only other
quantity that determines the correct digit. For each block $\ell$ and
position $p$, we replace the target's $h^{(\ell)}(p)$ with that of its
matched source example, let the forward pass continue, and measure
\begin{equation}
  \accsrc = \frac{1}{N} \sum_{i=1}^{N}
  \mathbf{1}\!\left[\hat d^{\,\text{patched}}_{c,i} =
  (S_{c,i} + c_s) \bmod 10\right],
\end{equation}
the fraction of examples whose digit becomes the one implied by the
source carry. Patching a position the model does not use leaves the
target's own digit, which differs from the source-carry value, so it
gives $\accsrc \approx 0$. Per-example matching matters: patching the unmatched
source-group mean instead conflates the carry with variation in $S_c$
and gives $\accsrc \approx 0.10$, against 0.85--0.95 with matching
(column~4, carry $5 \to 10$, all layers patched).

Patching at the \emph{last} block is a sanity check rather than
evidence: after block 5, the prediction slot's state alone determines
the output, so replacing it transfers the matched source's answer by
construction. Evidence about where the carry is held comes from
blocks $\le 4$, where later layers can still read other positions and
overwrite the patched value.

% =====================================================
\section{Results}
\label{presentation}

\subsection{Separators make multiplication learnable}
\label{sec:format}

\Cref{tab:format} compares the two formats. Without separators, the
model reaches 1\% exact match; with them, 89\%. Tracking per-position
error during training (\cref{app:dynamics}) shows that in the compact
format the middle positions never converge, while in the spaced
format all positions converge, from the outside in.  A plausible reason why separators help is that
they give each digit a prediction slot with no other job; in the
compact format, the same hidden state must both represent a digit and
predict the next one.

\begin{table}[h]
  \centering
  \small
  \begin{tabular}{lc}
    \toprule
    Format & Best acc.\ (\%) \\
    \midrule
    spaced  & 89 \\
    compact & 1 \\
    \bottomrule
  \end{tabular}
  \caption{Answer-only $4 \times 4$ multiplication, zero-padded,
  trained until convergence on 1M examples (best of three seeds for spaced; a single run for compact).}
  \label{tab:format}
\end{table}

\subsection{Token embeddings: digits on a circle, space at the centre}
\label{sec:tokenemb}

\begin{figure}[h]
  \centering
  \includegraphics[width=0.95\columnwidth]{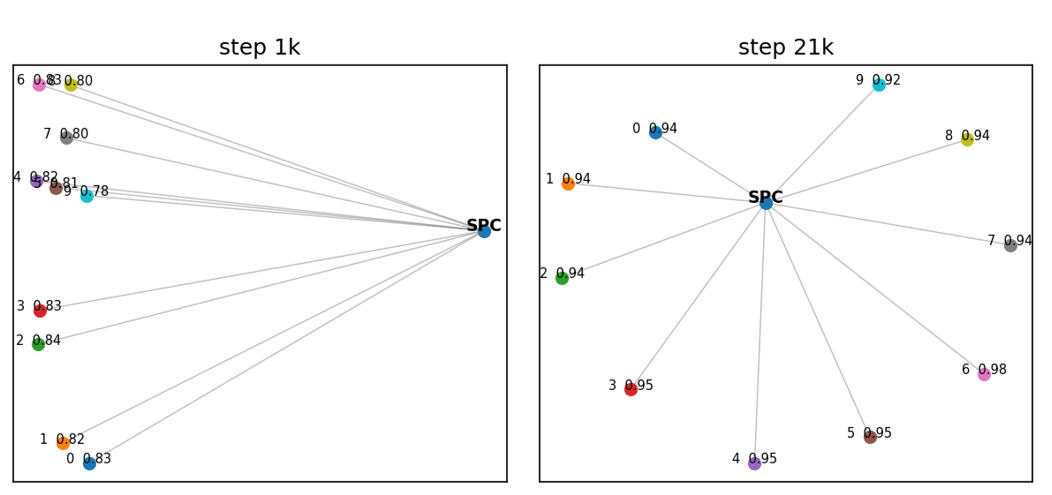}
  \caption{PCA of the input token embeddings at training steps 1k
  (left) and 21k (right). After training, digits 0--9 form a circle
  ordered by value; the space token sits near the centre.}
  \label{fig:pca10_tokenembedding}
\end{figure}

\Cref{fig:pca10_tokenembedding} shows PCA of the input token
embeddings: after training, digits 0--9 lie on a circle ordered by
value, the Fourier-style encoding documented in prior work
\citep{nanda2023progress, kantamneni2025trigonometry,
bai2025multiplication}; the space token sits at the centre, as an
input it carries no digit value.

\subsection{Per-position learning dynamics}
\label{app:dynamics}

\begin{figure}[h]
  \centering
  \includegraphics[width=0.5\textwidth]{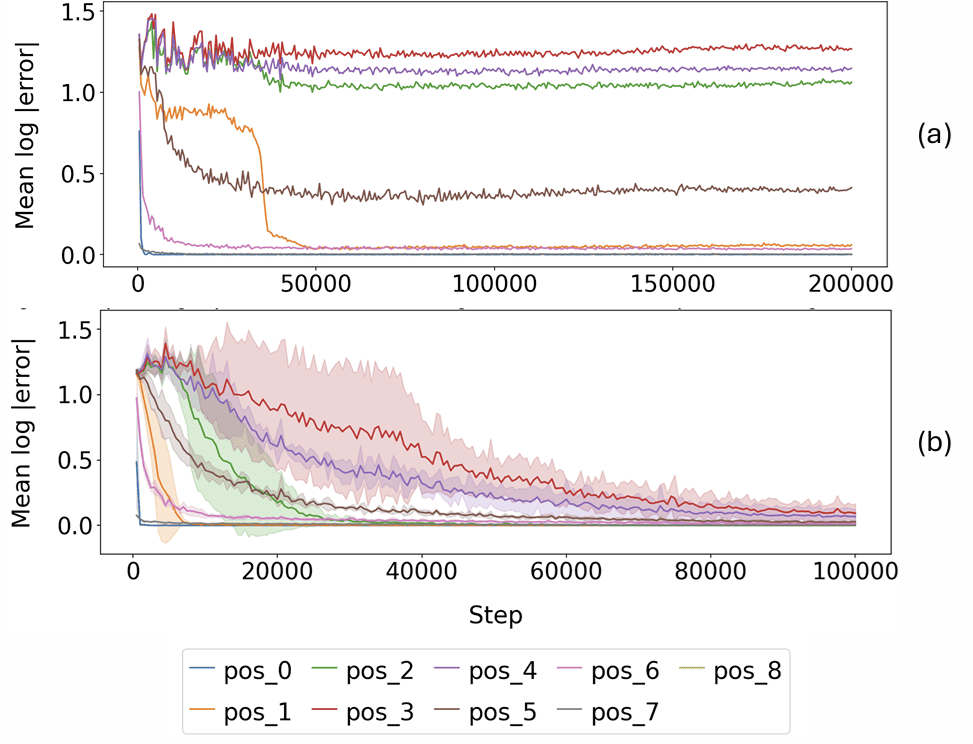}
  \caption{(a): Per-position log error over 200K steps in the compact format. (b):Per-position log error over 100K steps in the spaced
  format, averaged over three seeds. Curves are output digit
  positions, LSB-first (pos~0 = units).  }
  \label{fig:per-position-err}
\end{figure}

For each output position $i \in \{0, \dots, 7\}$, indexed LSB-first,
we measure the per-position log error $\log(|p_i - g_i| + 1)$ on the
held-out set throughout training (\cref{fig:per-position-err}). In the
spaced format, positions converge in a consistent order across seeds:
the outer positions (0, 7) first, then 1 and 6, then 2 and 5, and the
middle positions (3, 4) last. In the compact format, the middle
positions never converge. A model at intermediate exact-match accuracy
is not ``halfway done'': it has solved several positions cleanly and
is still struggling on a few.

\begin{figure}[h]
  \centering
  \includegraphics[width=0.4\textwidth]{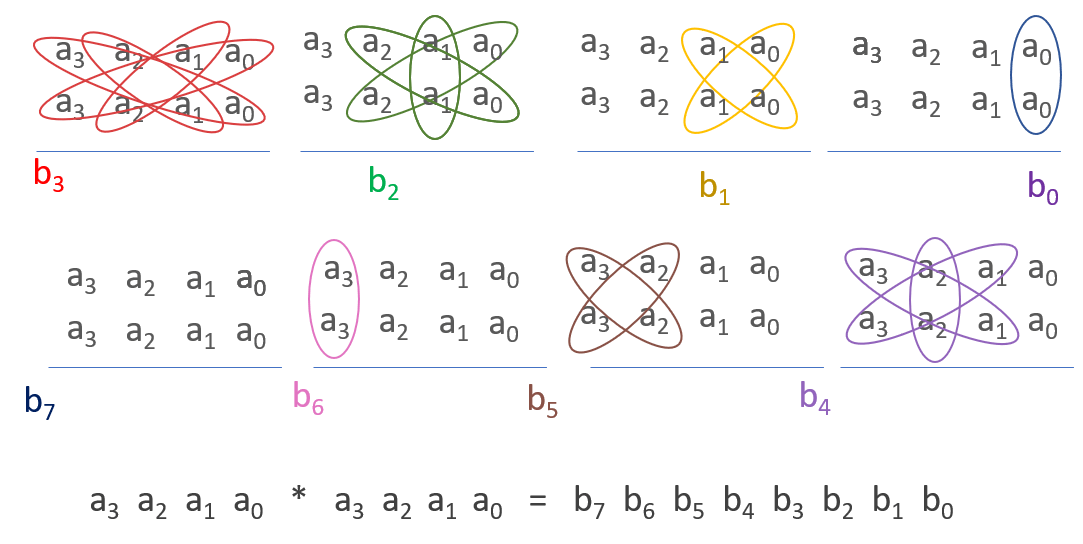}
  \caption{Direct input-digit dependencies for each output position
  in $4 \times 4$ multiplication, ignoring carry propagation. Outer
  positions (0, 7) depend on a single input-digit pair, while middle
  positions (3, 4) depend on all four pairs in their column.}
  \label{fig:digit-dependencies}
\end{figure}

This order matches the structure of long multiplication
(\cref{fig:digit-dependencies}): positions 0 and 7 depend on a single
input-digit pair, positions 1 and 6 on two, and position 3  on
all four pairs in their column plus the largest carries. The same
positions whose prediction slots cover the most of their ring (3 and
4, \cref{sec:carry_geometry}) are those learned last.

\subsection{The angle on  the ring is the carry-in}
\label{sec:carry_geometry}

\begin{figure}[h]
  \centering
  \includegraphics[width=0.95\columnwidth]{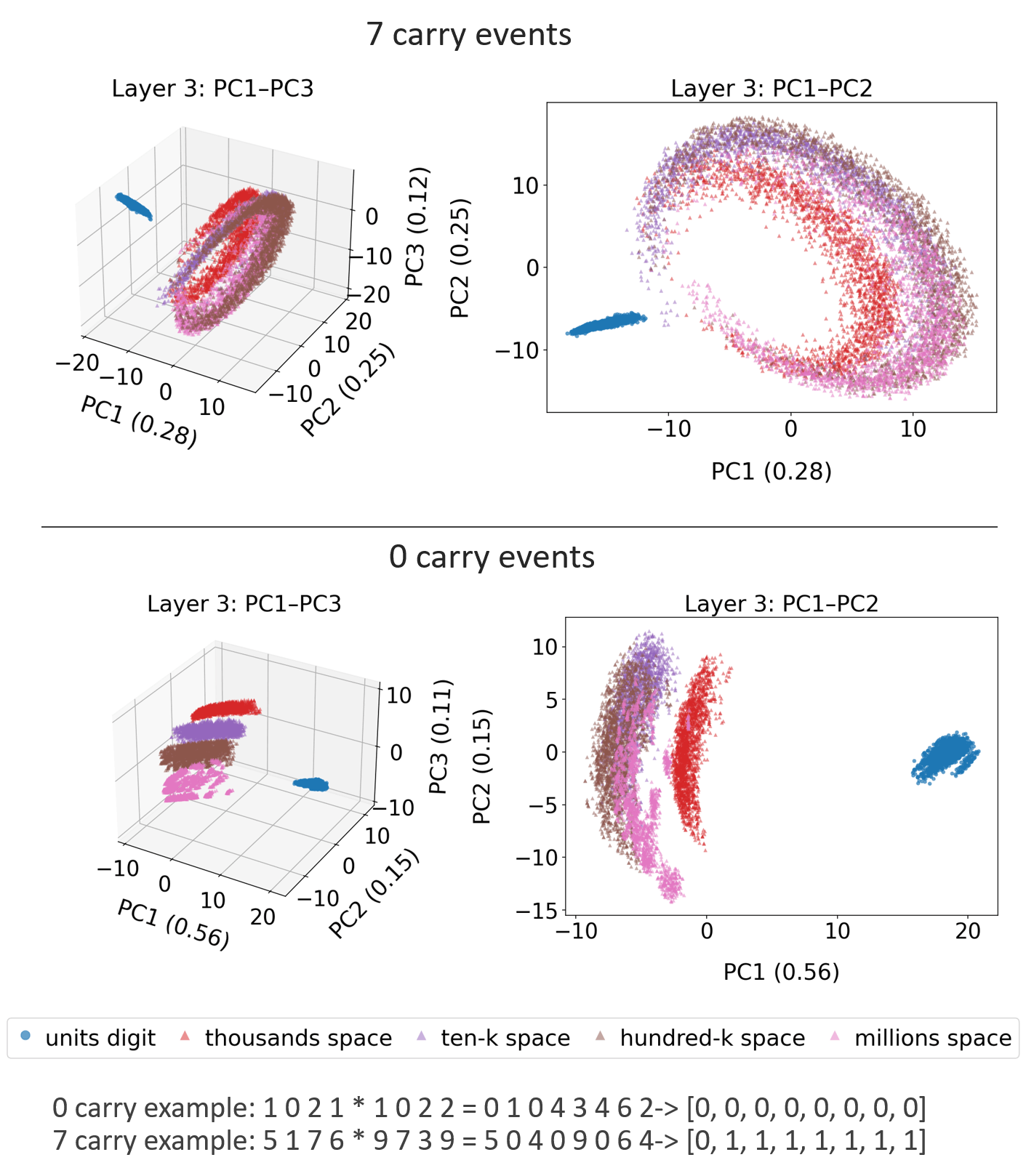}
  \caption{Residual stream at block 3 for examples with $k = 0$
  (e.g.\ \texttt{1021 * 1022}) and $k = 7$
  (e.g.\ \texttt{5176 * 9739}). The same prediction slots form arcs
  when every carry-in is zero and nearly closed rings when carry-ins
  vary widely.}
  \label{fig:residuals_carries}
\end{figure}

On the answer side, the space tokens stop being neutral: separated by
output column, each prediction slot forms its own ring-shaped cloud
in the leading principal components. Splitting examples by carry
count $k$ and fitting one PCA per group,
\cref{fig:residuals_carries} shows the two extremes: at $k = 0$ the
prediction slots form short arcs, at $k = 7$ nearly closed rings, and
intermediate $k$ interpolate (\cref{fig:mainpicture}).

\begin{figure}[h]
  \centering  \includegraphics[width=1.0\columnwidth]{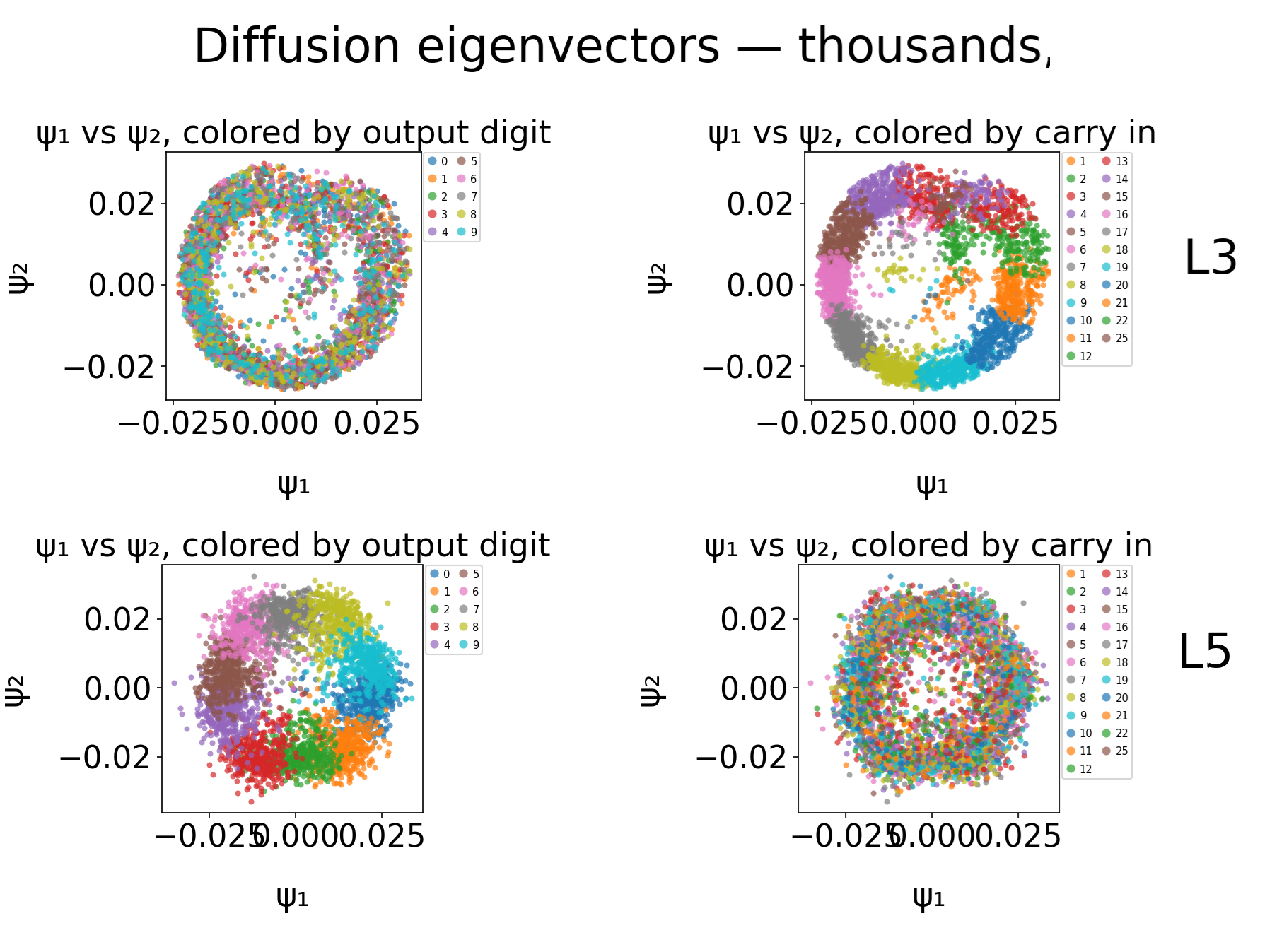}
  \caption{Diffusion map ($\psi_1$ vs $\psi_2$) of the column-3
prediction slot after blocks 3 (top) and 5 (bottom), for 7-event carry examples, on seed 10. Left: colored by output digit 
  (0--9). Right: colored by carry-in (0--30). After block 3, the
  carry-in is encoded by the angle on the ring decodable by the carry-in amount. At block 5 the ring of carry-in
  dissolves and the shape becomes decodable by digit, as the slot turns the carry
  into the output digit.}
  \label{fig:diffusion_map}
\end{figure}

The diffusion map of the column-3 prediction slot
(\cref{fig:diffusion_map}) shows what the ring encodes: at block 3,
points are ordered around it by their carry-in $\gamma_3$. This
explains the dependence on $k$. Examples with many carries take many
different carry values and so fill a larger part of the ring;
examples with no carry all share $\gamma = 0$ and sit on a small part
of it. The ring is a \emph{carry dial}, and the carry count decides
how much of the dial is used. At block 5 the ring points are no longer ordered by carry-in but group by output digit instead, as the slot commits to the
digit it will emit.

\subsection{The carry-in is causally used at the prediction slot}
\label{sec:causal}

\begin{figure}[h]
  \centering
  \includegraphics[width=1.2\columnwidth]{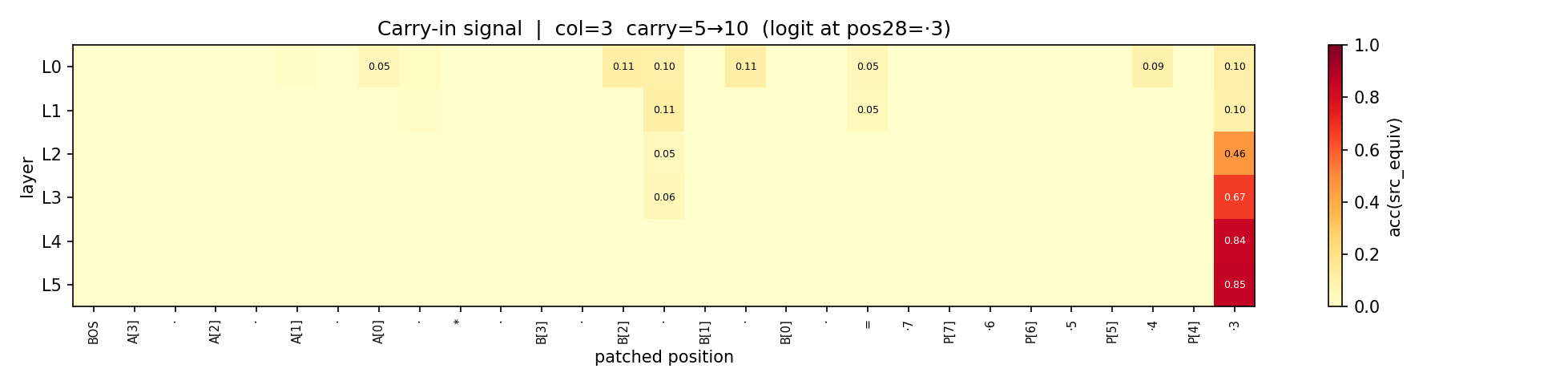}
  \caption{Matched activation patching for output column~3 (target
  carry-in 10, source carry-in 5). Each cell is the fraction of target
  examples whose digit changes to the source-carry value when the
  hidden state at that (block, position) is replaced by that of the
  matched source example. In blocks 0–1, patching operand and answer positions changes the column-3 digit only weakly ($\le 0.11$). From block 2, the effect concentrates at the column-3 prediction slot. The block-5 row is a sanity
  check (see \cref{sec:setup}).}
  \label{fig:patching_heatmap}
\end{figure}

The ring geometry shows a correlation between prediction-slot states
and the carry-in. To test whether the model uses these states, we
apply matched patching (\cref{sec:setup}) to output column~3, with
source carry-in 5 and target carry-in 10
(\cref{fig:patching_heatmap}). In blocks 0--1 the effect is small
(0.05--0.11) and spread over input digits and answer positions.
From block 2 it concentrates on the prediction slot of column~3:
$\accsrc$ reaches 0.46, 0.67 and 0.84 after blocks 2, 3 and 4, with
every other position near zero. Replacing this single vector at an
intermediate layer, while later layers can still read the rest of
the sequence, is enough to make the model answer as if the carry were
different.

\begin{figure}[h]
  \centering
  \includegraphics[width=1.052\columnwidth]{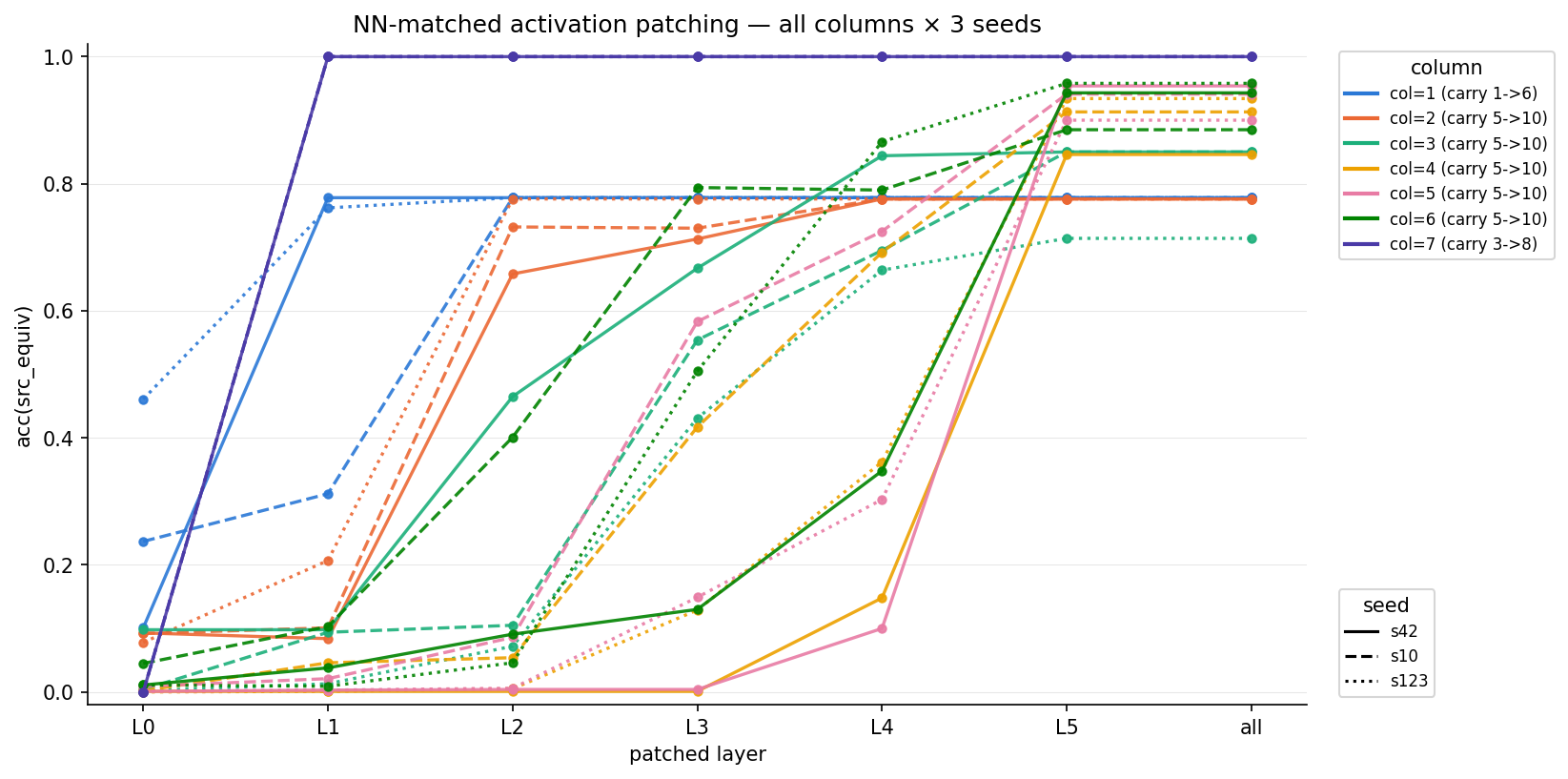}
  \caption{Matched patching of the prediction slot of each output
  column (colors, columns 1--7), one block at a time (x-axis; ``all'':
  every block at once, which includes the last block). Line style:
  seed. Values at block 5 and ``all'' are sanity checks.}
  \label{fig:patching_heatmap_combined}
\end{figure}

We repeat layer-by-layer patching of each column's prediction slot
for columns 1--7 and three seeds
(\cref{fig:patching_heatmap_combined}). Two boundary columns behave
differently from the interior ones. Column~1, whose carry-in depends
on a single upstream product, and column~7, which has no partial
product of its own ($d_7 = \gamma_7$), reach their plateau (0.78 and
1.00) by block 1 in all three seeds. For the interior columns 2--6,
whose carry-in accumulates a chain of upstream columns, the effect
rises only in later blocks, with an onset that varies modestly across
seeds.

\begin{figure}[H]
  \centering
  \includegraphics[width=0.85\linewidth]{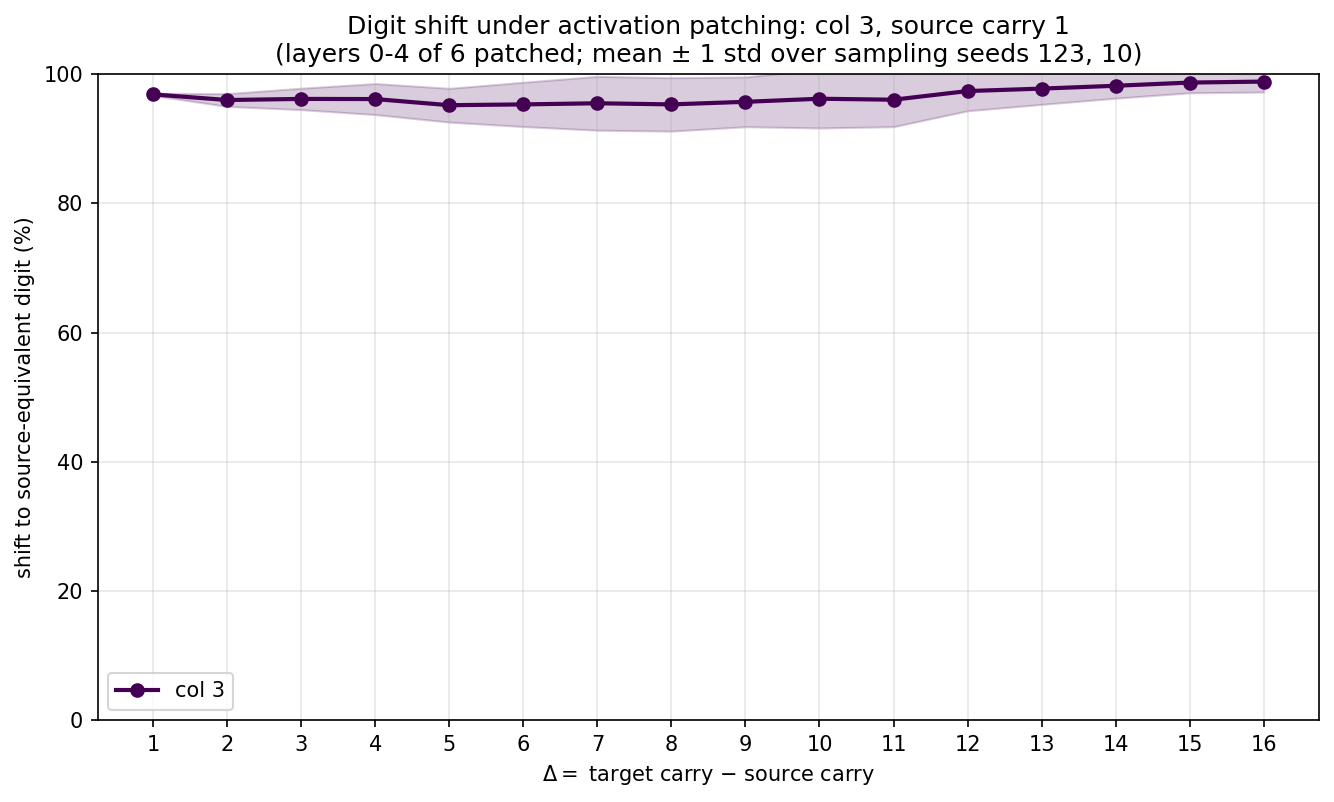}
  \caption{Fraction of patched examples whose digit follows
$(d^{\text{before}}-\Delta)\bmod 10$, for several carry differences
$\Delta$, averaged over two model seeds (10 and 123). The separator
tokens just before  column-3 digit is patched at every block except
the last (blocks 0--4); the last block and all other token positions
keep the target's states. Matched patching; row L5 is a sanity check}
  \label{fig:delta_carry_col3}
\end{figure}

To test whether the slot carries the carry \emph{value}, we vary the
carry difference $\Delta=\gamma^{\text{target}}-\gamma^{\text{source}}$
and patch only the prediction slot up to block 4 (equivalent to
patching it at block 4), leaving the last block free to read the rest
of the sequence. If the slot holds the carry, the digit should become
$(d^{\text{before}}-\Delta)\bmod 10$ by \cref{eq:carry}. For column 3
it does in 95--100\% of cases for every $\Delta$ tested
(\cref{fig:delta_carry_col3}), as for column 1,2, 7
(\cref{fig:delta_carry}). For columns 4--6 the shift is only
partial, consistent with their carry still being held at the
neighboring slot at block 4. At $\Delta = 10$ the target digit is
unchanged, so this point does not discriminate. Because pairs are
matched on $S_c$, a pass is also consistent with the slot holding the
column result $S_c + \gamma_c$ rather than $\gamma_c$ alone.

The route to the prediction slot differs between columns. For
column~4 (\cref{app:relay}), patching the column's own prediction slot
has no effect up to block 3; at blocks 2--3 the effect sits
instead at the prediction slot of the preceding, higher-order column
($\accsrc = 0.36$--$0.39$), which also needs $\gamma_4$ to predict
its own digit, before moving to the column-4 slot (0.15 after block
4). The carry is thus assembled at the neighboring answer slot and
relayed to its own slot, rather than accumulated in place.

\subsection{Errors are off by one}
\label{sec:graceful}

\begin{figure}[!t]
  \centering
  \includegraphics[width=0.85\linewidth]{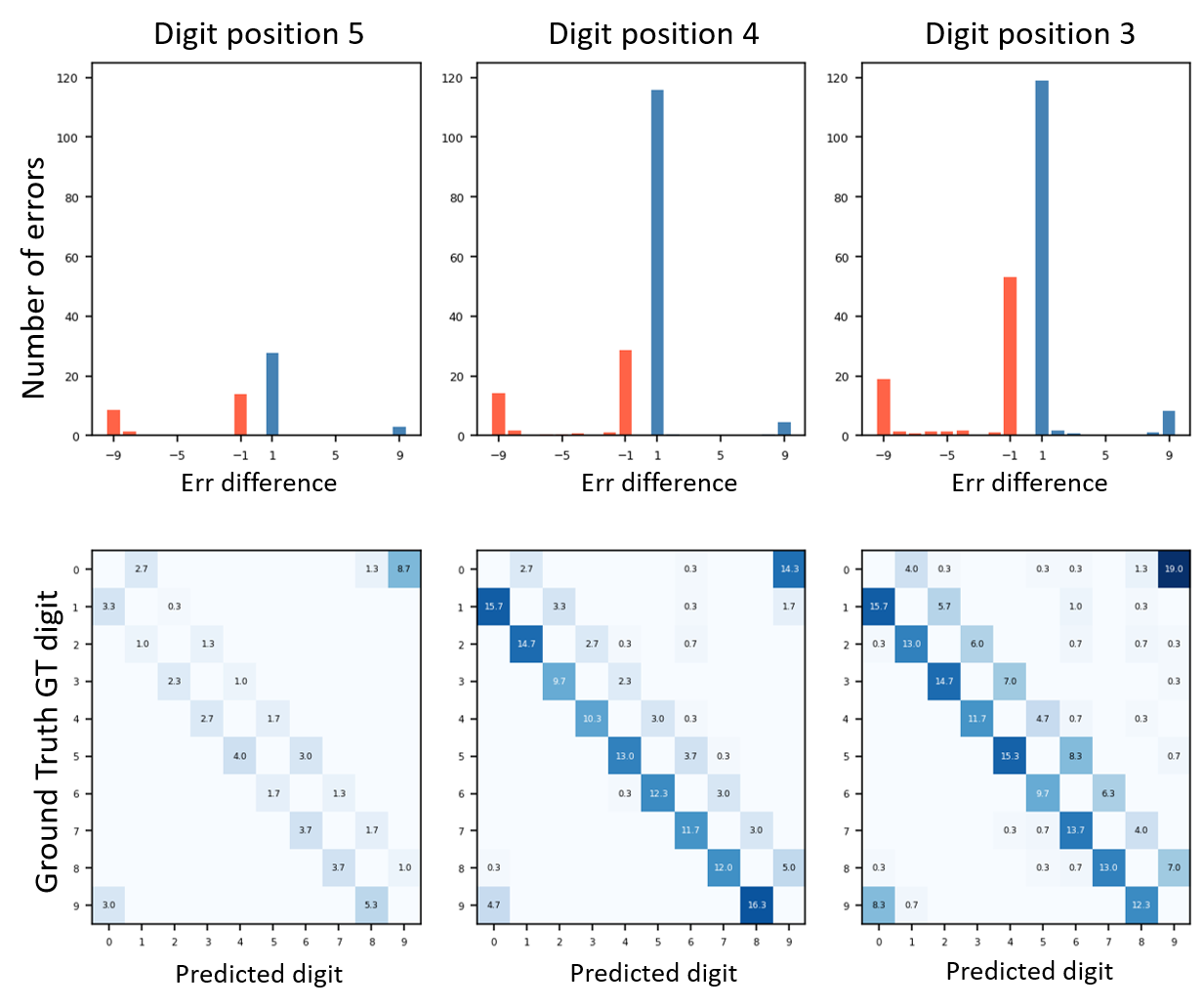}
  \caption{Distribution of prediction errors (prediction $-$ ground
  truth) at digit positions 5, 4 and 3 (LSB-indexed), with the
  prediction-vs-ground-truth confusion matrices. Positions 3 and 4
  accumulate the most errors, almost always off by $\pm 1$.}
  \label{fig:perpos_err}
\end{figure}

The errors that remain at convergence concentrate at positions 3 and
4 and are almost always off by $\pm 1$ (\cref{fig:perpos_err}). Two
readings are possible, and both point to the carry. By
\cref{eq:carry}, a carry estimated one unit too high or too low
shifts the digit by exactly one. And since digits are represented on
a circle (\cref{sec:tokenemb}), a small angular error also yields a
neighboring digit.

% =====================================================
\section{Conclusions}

Separators make carry computation learnable: with a space token
between digits, a small transformer reaches 89\% on answer-only
$4 \times 4$ multiplication; without it, 1\%. In the models that learn
the task, the carry-in of each column is encoded as an angle on a ring
at the position that predicts the digit, and examples with more
distinct carries fill more of the ring. Matched activation patching
shows that this state is causally used at intermediate layers and
that for columns 4 and 5 the carry is first assembled at the neighboring
answer slot before reaching its own. These results complement, rather
than contradict, the ``bag of heuristics'' found in pretrained LLMs
\citep{nikankin2024arithmetic}, which were never trained on a format
that gives the carry a dedicated slot.

\paragraph{Limitations.}
 The main spaced configuration is trained with three seeds, but the
compact configuration uses a single run, and we use a single model
size and architecture. All results concern
$4 \times 4$ multiplication with one tokenization. The geometric
analysis uses one seed per figure, and the dependence of the ring on
the carry count $k$ reflects how many distinct carry values a group
covers, not a change in the encoding itself. On the causal side,
patching measures whether a single site is \emph{sufficient} to change
the digit, not whether information is present there: for columns 4--6
the carry is represented at the prediction slot but patching it alone
transfers it only partially. Because pairs are matched on $S_c$, a
successful patch cannot separate the carry $\gamma_c$ from the column
result $S_c + \gamma_c$; pairs with $S_c$ differing modulo 10 would.
Finally, patching locates where the carry is held, not how it is
computed: the attention heads that perform the relay are not
identified.

% =====================================================
\section{Future work}
\label{sec:future}

First, testing \emph{why} separators help: spaces only in the answer
versus only in the operands, a different filler token, and multiple
seeds per format at matched compute would separate the ``free
prediction slot'' explanation from extra computation positions or
column alignment. Second, tracking the ring during training alongside
per-position error would show whether the carry dial forms before the
hard columns are solved. Third, attention-head-level analysis of the
relay from the neighboring slot to the prediction slot would connect
the representational findings to a circuit-level account.

\bibliography{example_paper}
\bibliographystyle{icml2026}

%%%%%%%%%%%%%%%%%%%%%%%%%%%%%%%%%%%%%%%%%%%%%%%%%%%%%%%%%%%%%%%%%%%%%%%%%%%%%%%
% APPENDIX
%%%%%%%%%%%%%%%%%%%%%%%%%%%%%%%%%%%%%%%%%%%%%%%%%%%%%%%%%%%%%%%%%%%%%%%%%%%%%%%
\newpage
\appendix
\onecolumn

% =====================================================

% =====================================================
\section{Additional figures}
\label{app:figures}

\subsection{Ring coverage varies with the column}
\label{sec:position_caveat}

\begin{figure}[h]
  \centering
  \includegraphics[width=1\textwidth]{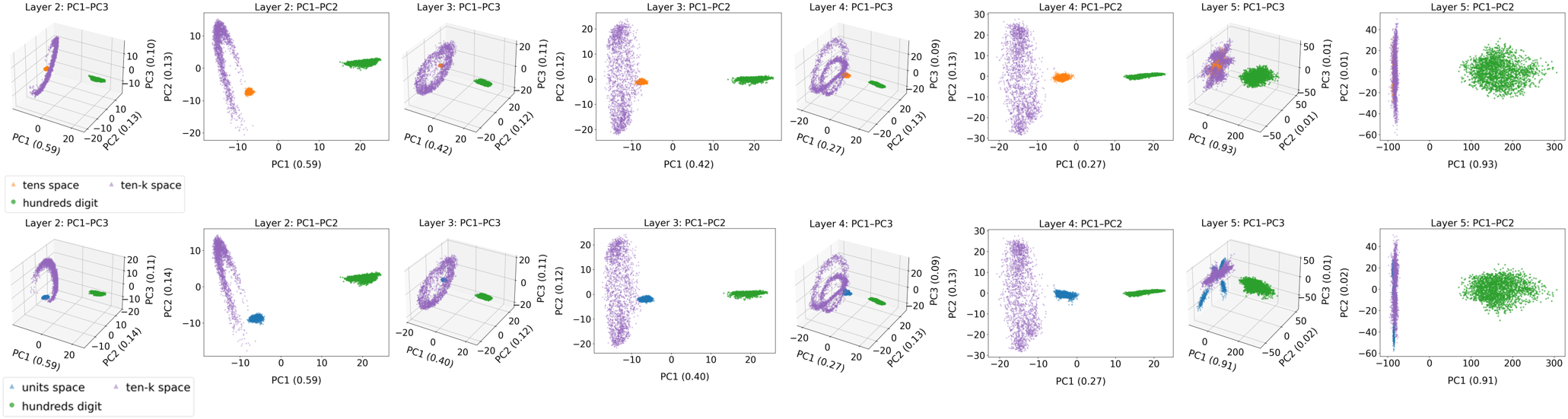}
  \caption{Prediction slots of positions 0 and 4 vs.\ 1 and 4 across
  layers. Position 4 forms a clear ring; positions 0 and 1 do not.}
  \label{fig:not_all_space_alike}
\end{figure}

The clearest rings appear at positions 3--7, whose carry-in takes a
wide range of values. Low-order positions (0--2), whose carry-in takes
only a few values, form arcs or partial structures even under heavy
carry load (\cref{fig:not_all_space_alike}), consistent with the
reading of the ring as a carry dial.

\subsection{Relay through the neighboring slot (different columns)}
\label{app:relay}

\begin{figure}[H]
  \centering
  \includegraphics[width=0.85\linewidth]{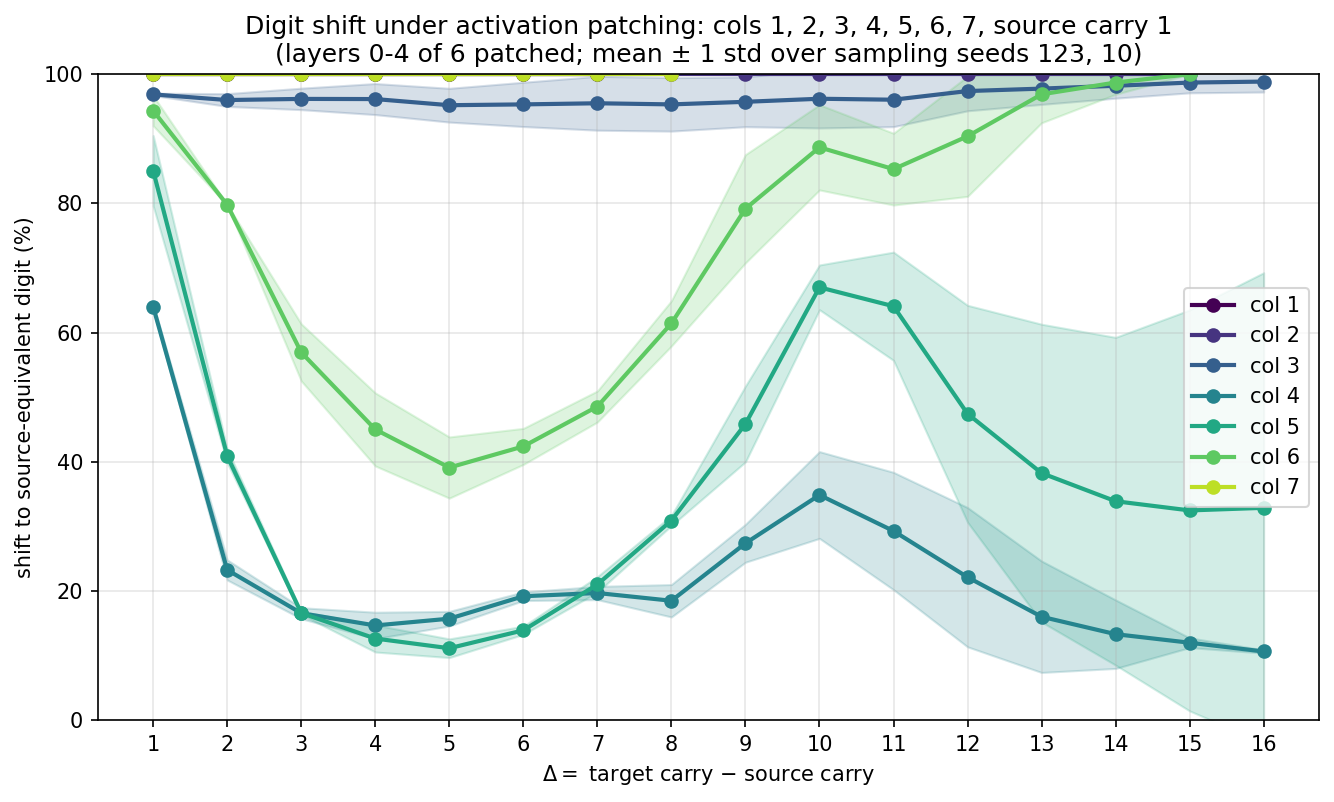}
  \caption{Fraction of patched examples whose digit follows
$(d^{\text{before}}-\Delta)\bmod 10$, for several carry differences
$\Delta$, averaged over two model seeds (10 and 123). The separator
tokens just before each column digit is patched at every block except
the last (blocks 0--4); the last block and all other token positions
keep the target's states. Matched patching; row L5 is a sanity check}
  \label{fig:delta_carry}
\end{figure}

\begin{figure}[h]
  \centering
  \includegraphics[width=0.9\textwidth]{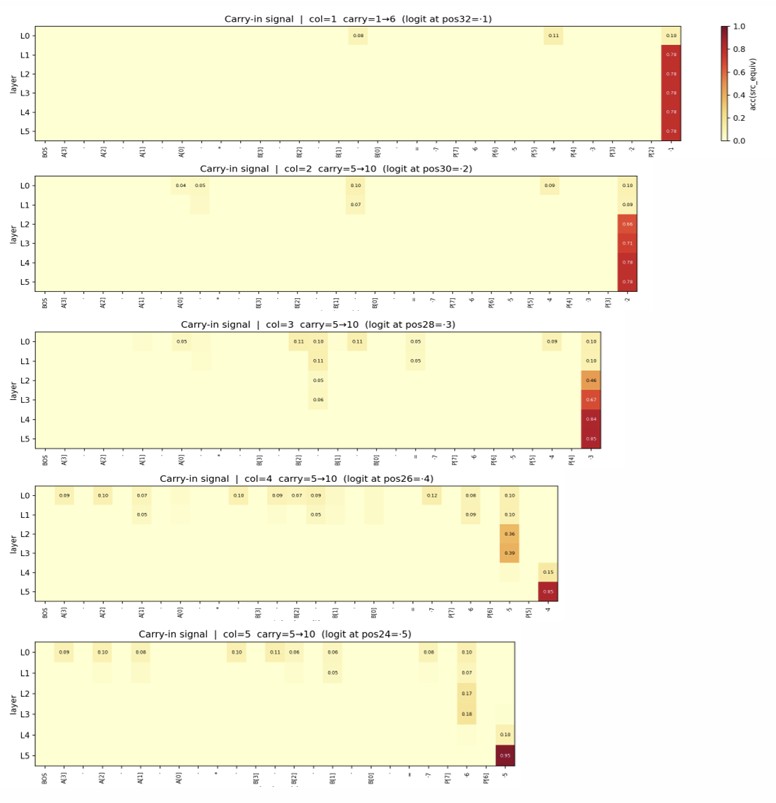}
  \caption{Activation patching for output columns $1$ to $5$. Each cell is the
  fraction of target examples whose digit changes to the source-carry value
  when the hidden state at that (layer, position) is replaced. Matched patching; row L5 is a sanity check}
  \label{fig:patching_col4}
\end{figure}

\end{document}